\documentclass[letterpaper, 10 pt, journal, twoside]{IEEEtran}

\IEEEoverridecommandlockouts                              

\usepackage{amssymb}
\usepackage{amsmath}
\usepackage{mathtools}
\usepackage{algorithm}
\usepackage{algpseudocode}
\usepackage{tikz}
\usepackage{layouts}
\usepackage{todonotes}
\usetikzlibrary{shapes.geometric, arrows}
\usepackage{xcolor}
\usepackage{url}
\usepackage{lipsum}
\usepackage{cite}
\usepackage{hyperref}

\usepackage{tensor}
\usepackage{stackengine}
\usepackage{hyperref,etoolbox}
\usepackage{fnpos}
\usepackage{fancyhdr}
\usepackage{pifont}
\usepackage{setspace}
\usepackage{multicol}
\usepackage{multirow}
\usepackage{graphicx}
\newtheorem{definition}{Definition}[section]

\makeatletter
\patchcmd{\ALG@step}{\addtocounter{ALG@line}{1}}{\refstepcounter{ALG@line}}{}{}
\newcommand{\ALG@lineautorefname}{Line}
\makeatother

\newcommand{\pnorm}[2]{\left\lVert#1\right\rVert_{#2}}

\newcommand{\Real}{\mathbb R}

\newcommand{\task}{\mathcal{T}}
\newcommand{\q}{\mathbf{q}}
\renewcommand{\v}{\mathbf{v}}
\newcommand{\x}{\mathbf{x}}
\newcommand{\xb}{\bar{\mathbf{x}}}

\newcommand{\txb}{\bar{\tilde{\x}}}
\renewcommand{\u}{\mathbf{u}}
\newcommand{\X}{\mathcal{X}}
\newcommand{\U}{\mathcal{U}}
\newcommand{\ub}{\bar{\mathbf{u}}}

\newcommand{\tub}{\bar{\tilde{\u}}}
\newcommand{\A}{\mathbf{A}}
\newcommand{\B}{\mathbf{B}}

\newcommand{\pt}{\mathbf{P}}     
\newcommand{\z}{\mathbf{z}}

\newcommand{\trans}[1]{#1^\mathrm{T}}
\newcommand{\spatial}[4]{\tensor*[^{#2}]{#1}{^{#3}_{#4}}}
\newcommand{\defeq}{\vcentcolon=}

\DeclareMathAlphabet\mathbfcal{OMS}{cmsy}{b}{n}

\DeclareMathOperator*{\lexmin}{lex\,min}
\DeclareMathOperator*{\lex}{lex}
\DeclareMathOperator*{\lexless}{<_{\lex}}

\definecolor{ashgrey}{rgb}{0.7, 0.75, 0.71}

\def\Vec#1{\!\!\hbox{$#1$\kern-0.38em\lower0.85em\hbox{$\vec{}\,$}}\,}%
\newcommand{\bbm}{\begin{bmatrix}}
	\newcommand{\ebm}{\end{bmatrix}}
\DeclareMathAlphabet{\mbf}{OT1}{ptm}{b}{n}

\usepackage{xcolor}
\newcommand{\revision}[1]{{\color{black} #1}}
\newcommand{\rerevision}[1]{{\color{black} #1}}
\newtheorem{preexample}{\sf\bfseries Example}

\newtheorem{prethm}{\sf\bfseries Theorem}

\newtheorem{prelem}{\sf\bfseries Lemma}

\newtheorem{preprop}{\sf\bfseries Proposition}

\author{Xintong Du, Siqi Zhou and Angela P. Schoellig%
\thanks{The authors are with the Learning Systems and Robotics Lab
(www.learnsyslab.org) at the Technical University of Munich, Germany, and
the University of Toronto Institute for Aerospace Studies, Canada. They are also affiliated with the University of Toronto Robotics Institute, the Munich Institute of Robotics and Machine Intelligence (MIRMI), and the Vector Institute for Artificial Intelligence. {\tt\small E-mail:  xintong.du@utias.utoronto.ca, siqi.zhou@robotics.utias.utoronto.ca, angela.schoellig@tum.de}}%

}

\title{\LARGE \bf
Hierarchical Task Model Predictive Control for Sequential Mobile Manipulation Tasks
}

\begin{document}

\maketitle

\begin{abstract}
Mobile manipulators are envisioned to serve more complex roles in people's everyday lives. With recent breakthroughs in large language models, task planners have become better at translating human verbal instructions into a sequence of tasks. However, there is still a need for a decision-making algorithm that can seamlessly interface with the high-level task planner to carry out the sequence of tasks efficiently. In this work, building on the idea of nonlinear lexicographic optimization, we propose a novel Hierarchical-Task Model Predictive Control framework that is able to complete sequential tasks with improved performance and reactivity by effectively leveraging the robot's redundancy. Compared to the state-of-the-art task-prioritized inverse kinematic control method, our approach has improved hierarchical trajectory tracking performance by $\mathbf{42\%}$ on average when facing task changes, robot singularity and reference variations.
\revision{\rerevision{Compared to a typical single-task architecture, our proposed hierarchical task control architecture enables the robot to traverse a shorter path in task space and achieves an execution time $\mathbf{2.3}$ times faster when executing a sequence of delivery tasks}. We demonstrated the results with real-world experiments on a 9 degrees of freedom mobile manipulator.}
\end{abstract}
\setlength{\textfloatsep}{10pt}

\begin{IEEEkeywords}
Mobile Manipulation; Redundant Robots; Whole-Body Motion Planning and Control
\end{IEEEkeywords}

\section{Introduction}
\begin{figure}[t!]
\centering
  \includegraphics[width=\linewidth,trim={0 3cm 0 1cm}, clip]{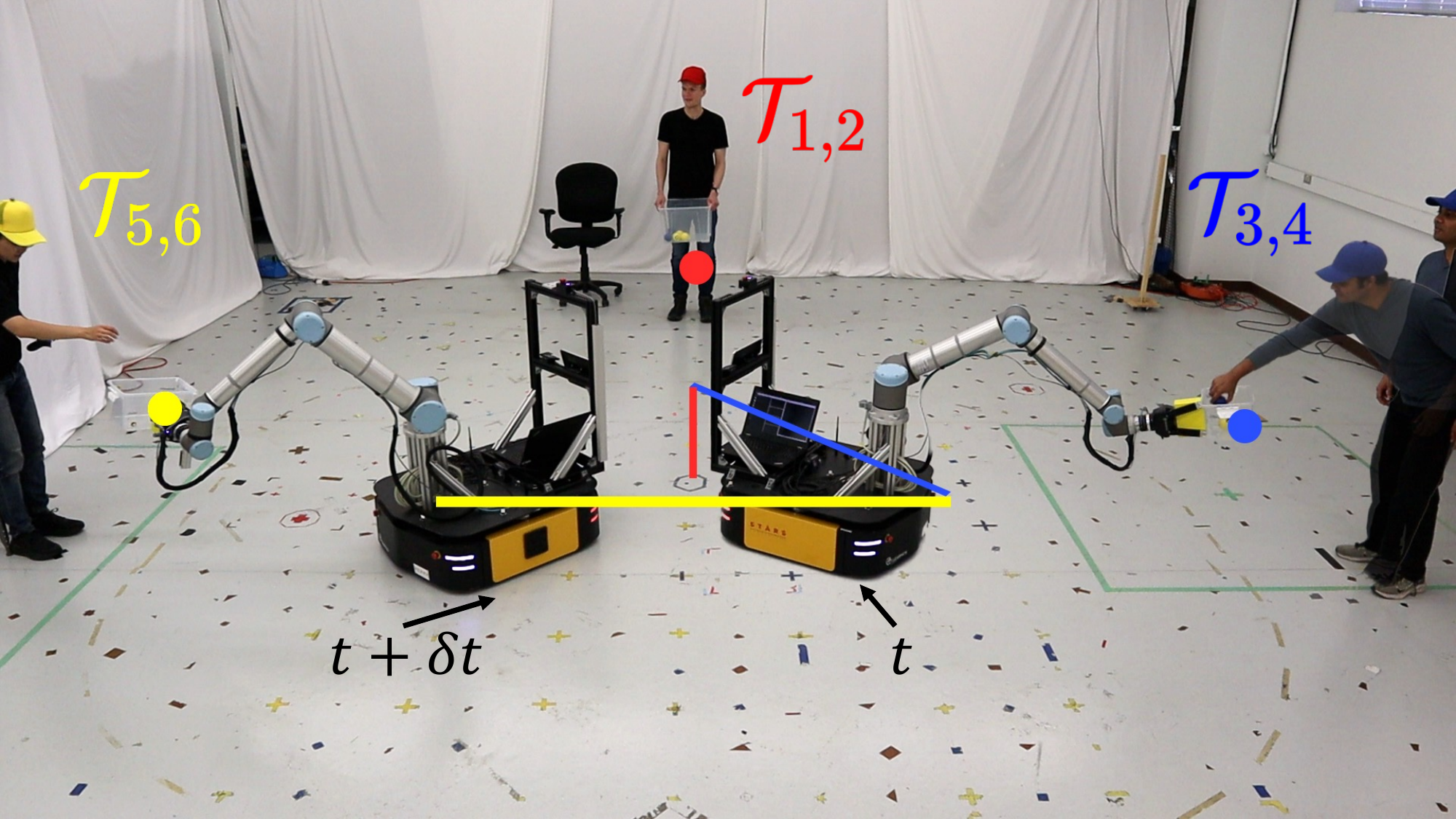}
  \caption{Our mobile manipulator executes a sequence of delivery tasks while leveraging redundancy for efficiency. Our robot needs to deliver to/pick up from three people following the order, {\color{red} Red},  {\color{blue} Blue}, {\color{yellow}Yellow}. The high-level goal is decomposed into a sequence of alternating base trajectories (strips) and end effector waypoints (dots). At time $t_0$, the robot leverages its redundancy to the current EE task $\mathcal{T}_4$ (Blue dot) to perform the subsequent base task $\mathcal{T}_5$ (Yellow strip) so that it can reach its next EE waypoint $\mathcal{T}_6$ faster. In this test, our proposed HTMPC motion control architecture is 2.3 times faster than the typical single-task method. A complete video can be found at \href{http://tiny.cc/htmpc}{http://tiny.cc/htmpc}.} \label{fig:main}
\end{figure}

Taking the form of a mobile base with robotic arms mounted on top, mobile manipulators have always been envisioned to serve more complex service roles in people's everyday lives. 
These roles require mobile manipulators to perform a sequence of manipulation tasks scattered at distant locations in a complex environment to fulfill high-level tasks instructed by a human \cite{ahn_as_2022}. We refer to this type of task as the sequential mobile manipulation task (see \autoref{fig:main}).

\begin{figure*}[ht!]
\centering
\includegraphics[width=\textwidth, height=3.5cm]{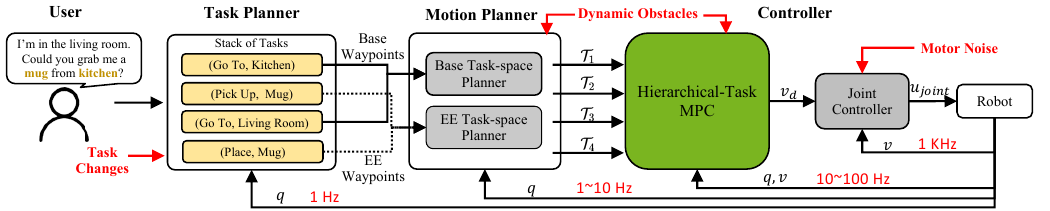}
  \caption{Proposed HTMPC planning and control architecture for sequential mobile manipulation tasks. Autonomy modules are represented as blocks. Arrows indicate the direction of information flow. Feedback loops from the robot are presented along with their desired close-loop frequency for reactive behaviours. External disturbances or changes coming from the robot or environment are specified as incoming arrows to each autonomy module affected.} \label{fig:arch}
\end{figure*}

Most control architectures for sequential mobile manipulation tasks \revision{consist of cascaded modules running at different frequencies} \autoref{fig:arch}. Given an instruction, task planners decompose it into a sequence of tasks to be executed by the \revision{subsequent} motion planning and control modules. With recent breakthroughs in large language models, task planners have become better at understanding human verbal instructions. However, the motion planning and control \revision{modules} could still be improved to enhance optimality and/or reactivity.

Typical motion planning and control methods for executing sequential tasks treat each task separately \revision{\cite{carius_deployment_2018, haviland_holistic_2022,pankert_perceptive_2020}}. As a result, robots can only execute one task after another and, for safety concerns, may need to come to a complete stop in between. Efficiency can be significantly improved if robots can attempt multiple tasks by leveraging their kinematic redundancy. For example, mobile manipulators should be able to perform an end effector (EE) task while moving its base towards its next goal. It should also be able to stop moving its base immediately before compromising the EE task. 
In addition to optimality and efficiency, it is also important to keep robots reactive to task changes or dynamic objects that are often seen in the real world.
However, most methods that leverage the robot's redundancy take a planning-oriented approach which puts a strong emphasis on optimality over reactivity \revision{\cite{tazaki_constraint-based_2014, zimmermann_go_2021,thakar_accounting_2019,thakar_manipulator_2022}}.

In control literature, leveraging the robot’s kinematic redundancy for executing ordered tasks is known as the hierarchical task control problem and has been widely studied in the past decades \cite{escande_hierarchical_2014}. In this work, our goal is to solve the sequential mobile manipulation problem using the hierarchical task control framework. We aim to solve the problem with low computational costs to maintain sufficient reactivity to real-world changes or disturbances. Although sequential mobile manipulation can involve a wide variety of tasks, similar to~ \cite{thakar_manipulator_2022, burgess-limerick_architecture_2023}, we focus on one typical example where human inputs are interpreted as a sequence of alternating base and EE waypoints by the task planner (\autoref{fig:arch}). However, it is worth mentioning that our proposed method does not assume a specific type of task sequence except that all tasks are tracking tasks.

Our key contributions are as follows:
\begin{enumerate}
    \item Formulate the sequential mobile manipulation control problem as a Hierarchical Task Model Predictive Control (HTMPC) problem and introduce a reformulation of the lexicographic optimality constraint in HTMPC to make it feasible to solve online.

    \item Propose a novel motion planning and control architecture 
    \revision{optimized for the HTMPC controller to} leverage the robot's kinematic redundancy for sequential tasks.
    
    
  \revision{
  \item Show that our proposed HTMPC outperforms the state-of-the-art inverse-kinematic-based method when facing task changes, robot singularity, and variations in trajectories. Show that the proposed hierarchical-task control architecture has improved \rerevision{efficiency and reactivity in robot behaviours for sequential tasks compared to existing control architectures}.

  \item Demonstrate the above results in experiments on a 9 degrees of freedom (DoF) mobile manipulator}
\end{enumerate}
\textit{Notation}: Subscript $k$ denotes the discrete time index over the prediction horizon. 
We use $\bar{\mathbf{p}}_{i:j}$ to denote the vertically concatenated vector of $(\mathbf{p}_i, \mathbf{p}_{i+1}, \dots, \mathbf{p}_j )$. The notation for state and control input sequence in MPC is $\xb_{0:N}$,$\ub_{0:N-1}$ where $N$ is the prediction window size. We will omit the subscripts in MPC state and control input sequence for simplicity and write them as $\xb, \ub$.

\section{Related Work}

\subsection{Sequential Mobile Manipulation Control Architecture} \label{sec:MMArch}
 

 

Typical motion control methods for mobile manipulators use a task-space planner to generate trajectories for each EE or base waypoint in the sequence \cite{carius_deployment_2018,burgess-limerick_architecture_2023}.
The EE and base task-space planners work independently; consequently, the resulting plans are not coordinated either in space or time. To avoid conflicting motions, these plans are usually executed sequentially by either decoupled arm and base controllers \cite{carius_deployment_2018} or whole-body controllers  \cite{pankert_perceptive_2020,haviland_holistic_2022}. 
Regardless of which type of closed-loop controller is used, these methods are single-task approaches in nature as each task-space plan is executed individually. 
\revision{In this work, we aim to improve} efficiency significantly by leveraging the robot's redundancy to achieve multiple tasks, i.e. redundancy resolution.

For sequential mobile manipulation, redundancy resolution means coordinating the arm and base toward multiple tasks in the sequence. This can be achieved by using a whole-body planner to generate optimal joint-space trajectories given a sequence of waypoints \cite{tazaki_constraint-based_2014, zimmermann_go_2021,thakar_accounting_2019,thakar_manipulator_2022}. However, whole-body planners can be expensive for high DoF robots; it is difficult to re-plan joint-space plans to be reactive in task space. \revision{In this work, we re-distributed the workload between the motion planner and controller by interfacing them at the task-space level to achieve reactive robot behaviours in task space.} In \cite{burgess-limerick_architecture_2023, burgess-limerick_enabling_2023}, a set of trajectory design strategies was proposed to generate an arm-base coordinated trajectory to enable "Manipulation on the Move" (MotM). Although kinematic consistency is not guaranteed, this strategy has been proven effective and reliable for sequential pick-and-place tasks. In this work, we took a more general approach, hierarchical-task control, which applies to all trajectory tracking tasks with guaranteed kinematic consistency by directly reasoning with robot kinematics. 

\subsection{Hierarchical Task Control}
Hierarchical task control leverages robot redundancy to achieve multiple control objectives with different priorities. Optimality for a task with a higher priority should not be compromised by lower-level tasks. As a type of multi-objective optimization problem, hierarchical task control can be solved using a scalarization approach where the objective is a weighted sum of the objectives from individual tasks \cite{sathya_weighted_2021}. Most methods for hierarchical control consider a strict hierarchy, which is also the focus of this work. However, hybrid methods also exist that can deal with both strict and softened hierarchy \cite{dehio_dynamically-consistent_2019}. 

Most works in the past are based on the Inverse Difference Kinematic  Control (IDKC) method, where tasks are formulated as linear feedback control laws. Therefore, the corresponding Hierarchical Task IDKC  (HTIDKC) problem is also linear. HTIDKC can be solved analytically by finding solutions in the null space of preceding tasks \cite{hanafusa_analysis_1981}, which was later generalized to handle any number of tasks using a recursive algorithm \cite{siciliano_general_1991}. Reformulated as an optimization problem, HTIDKC can also be solved with a sequence of quadratic programs (QP) in \cite{kanoun_kinematic_2011}. In \cite{escande_hierarchical_2014}, a more scalable solver was proposed, which is the state-of-the-art Hierarchical Quadratic Programming (HQP) approach. 

The optimization formulation also inspires a few works on solving nonlinear hierarchical-task problems. In \cite{tazaki_constraint-based_2014}, hierarchical trajectory generation tasks are formulated as a nonlinear lexicographic optimization problem. 
In \cite{lee_mpc-based_2020}, an MPC-based hierarchical task space control method was proposed for an under-actuated and constrained robot. In \cite{romano_prioritized_2015}, dynamic programming approaches were extended to address optimal control problems with hierarchical objectives. All proposed methods were either demonstrated in simulation or on a low DoF robot. Their run-time efficiency is still yet to be demonstrated as a reactive controller on high DoF robots.

In this work, we focus on solving the nonlinear HTMPC problem for sequential tasks online to enable reactive behaviours for high-dimensional robots. In \cite{lee_real-time_2022}, a real-time MPC was proposed for an industrial manipulator performing hierarchical tracking tasks. However, the core hierarchical task control problem is tackled outside the MPC controller using the HTIDKC approach. In contrast, our work aims to solve the problem directly within the MPC framework.
\section{Preliminaries on Lexicographic Optimization}
We give a brief introduction to lexicographic optimization. We refer readers to 
\cite{ehrgott_multicriteria_2005}\label{sec:lexi-opt} for a complete treatment of this subject.
A lexicographic optimization problem can be written as follows:
\begin{equation}\label{def:lex-opt-def}
    \lexmin_{\x \in \X}\: [f_1(\x), \cdots, f_L(\x)],
\end{equation}
where $f_1,\dots, f_L$ is a list of tasks given in the desired order of decreasing priority. The goal is to find the optimal $\x \in \X$ that minimizes the vector objective in the sense of lexicographic order.
Lexicographic order between two vectors is defined as 
\begin{definition}\label{def:lex-ineq-def}
    For two vectors $\z^1, \z^2\in \Real^n$, $\z^1 <_{\lex} \z^2$ if and only if $\exists l^*\defeq \min\{l: z^1_l \neq z^2_l\}$, $z^1_{l^*} < z^2_{l^*}$. 
\end{definition}
Lexicographic optimality is defined as follows:
\begin{definition}\label{def:lex-opt-opt-def}
A feasible solution $\hat\x \in \X$ is lexicographically optimal or a lexicographic solution if there is no other $\x \in \X$ such that $\mathbf{f}(\x) \lexless \mathbf{f}(\hat \x)$.
\end{definition}
Local lexicographic optimality definition can be similarly formulated by limiting the set $\{\x \in \X\}$ to a small norm ball centered at $\hat\x$, $\{\x \in \X: \pnorm{\x - \hat{\x}}{} \leqslant \sigma \}$.
Intuitively, the lexicographic order of two vectors depends on the first index $l^*$ where the two vectors differ. The vector that has a smaller $l^*$th entry is also smaller in the sense of lexicographic order. 

Lexicographic order implements a sense of hierarchy, decreasing in priority from $z_0$ to $z_n$ since the order of $(z^1_l, z^2_l)$ becomes relevant only if all the preceding entries are \revision{equal}. Similarly, in~\eqref{def:lex-opt-def}, the order of scalar objectives in the vector also indicates their hierarchy in decreasing order from $f_1$ to $f_L$, so the optimality of $f_l$ is considered only after the optimality of $[f_1, \dots, f_{l-1}]$ have been established.


The interpretation of lexicographic order as a hierarchy gives rise to a common algorithm (\autoref{alg:Lex-Opt}) for solving lexicographic optimization problems \cite{ehrgott_multicriteria_2005}. 
Scalar objective functions are optimized sequentially from $f_1$ to $f_L$ as a single objective optimization problem \eqref{eq:lex-opt-st}. For each iteration, a new constraint is added \eqref{eq:lex-opt-st-lexopt-constraint} to enforce lexicographic order by 
prohibiting the optimality of preceding objectives from being jeopardized while optimizing the current objective.


There are two challenges when applying \autoref{alg:Lex-Opt} to mobile manipulators: \textit{(i)} problem \eqref{eq:lex-opt-st} is nonlinear and non-convex, \textit{(ii)} with gradient-based solvers, it is costly to determine solution uniqueness. In this work, we focus on addressing \textit{(i)} to enable reactive behaviours for mobile manipulators executing sequential tasks while retaining local optimality. 
\revision{Moreover, we do not attempt to check the solution uniqueness for early termination; we will allow the algorithm to iterate through all tasks in the objective, although the solution might not be updated. To reduce redundant computation, we will manually choose the number of tasks so that their total dimension is smaller than the robot's DoF. }

\begin{algorithm}[t!] 
\caption{Lexicographic Optimization}\label{alg:Lex-Opt}
\begin{algorithmic}[1]
\Require \textit{Feasible set} $\X$ \textit{and objective function} $f$
\While{$l \leqslant L$}
\State $\textit{Solve the single objective optimization problem}$ 
\begin{subequations} \label{eq:lex-opt-st}
\begin{align}
    f^*_l(x) = \min_{x\in\X}\: & f_l(x)\\
    s.t.\:& f_i(x) = f^*_i(x), \label{eq:lex-opt-st-lexopt-constraint}\\
    &\; \text{for } i=1, \cdots, l-1
\end{align}
\end{subequations}
\State \textit{If \eqref{eq:lex-opt-st} has a unique optimal solution $\hat x_l$, STOP, $\hat x_l$ is the unique optimal solution}\label{alg:Lex-opt-line-uniqueness-test}
\State \textit{If \eqref{eq:lex-opt-st} is unbounded, STOP, \eqref{def:lex-opt-def} is unbounded.}
\State \textit{If $l=L$, STOP, the optimal solutions is.}
\begin{equation}
    \Big\{ x \in \X :f_l(x) = f^*_l(x), l=1, \cdots, L\Big\}
\end{equation}
\State $l \gets l+1$
\EndWhile
\State \Return \textit{Set of lexicographically optimal solutions}
\end{algorithmic}
\end{algorithm}
\section{Methodology}\label{sec:method}
Our proposed HTMPC architecture is presented in \autoref{fig:arch}. Similar to typical methods, the motion planner uses separate EE and base planners and generates a sequence of task-space trajectory tracking task \revision{$[\task_1, \task_2, \task_3, \task_4, ... ]$}. \revision{At any time step, only a part of this list is given to the HTMPC module, $[\task_{l_o},  \cdots,\task_{l_o+L-1}]$, which will be updated if $\task_{l_o}$ is completed by incrementing $l_o$ by 1}. HTMPC controller \revision{maintains the time-ordered sequence of the given tasks} and generates joint-space trajectories that coordinate the robot's all body parts toward multiple tasks. In this section, we present the HTMPC controller and show how it enables optimal and reactive behaviours for sequential mobile manipulation tasks.
\subsection{System and Task Model}
We consider a mobile manipulator with state $\x = \trans{[\trans{\q} \; \trans{\v}]}$ where $\q \in \Real^n$ denotes the generalized coordinates \revision{of both the base and the arm}, and $\v \in \Real^m$ denotes the generalized velocity. The generalized coordinate and velocity are related by $\dot \q = \mathbf{G}(\q) \v$ \revision{where $\mathbf{G}(\q)$ is a block diagonal matrix with two components: an identity matrix for the arm and a position-dependent matrix for the mobile base \cite{siciliano-robotics-mobile-modeling}.} We choose accelerations as the control inputs, $\u = \dot \v$. The feasible sets for robot state and control inputs are $\X$ and $\U$, respectively. Robot's kinematic model can be written as
\begin{equation}\label{eq:kinematics}
\dot \x = \A(\q)\x + \B\u.
\end{equation}
where 
\begin{equation}
   \A = \begin{bmatrix}
   \mathbf{0}_{n\times n}& \mathbf{G}(\q)\\
   \mathbf{0}_{m\times n}&\mathbf{0}_{m\times m}
   \end{bmatrix},\;
    \B = \begin{bmatrix}
   \mathbf{0}_{n\times m}\\
   \mathbf{I}_{m\times m}
   \end{bmatrix}.
\end{equation}

Let point A be a point rigidly attached to the robot. Following notations in\cite{tedrake_manipulation_textbook_2022}, its location as seen in the world frame can be determined using the robot's forward kinematics,
$
    \spatial{\pt}{}{A}{} =  \spatial{\mathbf{f}}{}{A}{}(\q)
$
where $\spatial{\mathbf{f}}{}{A}{}(\q): \Real^n \rightarrow \Real^3$ is the forward kinematics function. 

For a trajectory tracking task $\task$, the robot needs to follow a desired reference signal $\mathbf{r}(t): [0, T] \rightarrow \Real^s$ with a point on its body where $t$ is the control time. We assume reference trajectories are specified in vector space, and the distance between $\pt$ and $\mathbf{r}$ is 
\begin{equation}\label{eq:tracking_task_distance}
    dist(\pt, \mathbf{r}) = \pnorm{\pt - \mathbf{r}}{} = \pnorm{\mathbf{f}(\q) - \mathbf{r}}{} \defeq \pnorm{\mathbf{e}}{}.
\end{equation}

\subsection{Hierarchical-Task MPC}
We now present our HTMPC controller for the sequential trajectory tracking problem. At each control time $t$, the controller solves a lexicographic optimization problem for a pair of optimal state and control sequence, $\bar{\x}^*$ and $\bar{\u}^*$, over a prediction horizon $N$ steps into the future at a discretization time step $\Delta t$. The lexicographic optimization problem is 
\begin{subequations}\label{eq:HT-MPC}
\begin{align}
      \lexmin_{\bar{\x}, \bar{\u}} \; &[\mathcal{J}_1, \mathcal{J}_2. \cdots, \mathcal{J}_L] \label{eq:HT-MPC-cost-vec}\\
 \text{s.t.} \; & \x_{k+1} = \A(\q_k) \x_k + \B\u_k & \label{eq:HT-MPC-mdl}\\
 & \mathbf{sd}(\x_k) \geqslant \pmb{\delta}_{safe} \label{eq:HT-MPC-collisioncst}\\
 & \x_{k} \in \mathcal{X}, \u_{k} \in \mathcal{U}\label{eq:HT-MPC-xucst}, \; \forall k=0,1,\cdots,N-1\\
 & \x_0 = \x_o\label{eq:HT-MPC-xocst}.
\end{align}
\end{subequations}
where $\x_o = \x(t)$ is the current state of the robot. State and control input at prediction step $k$ are $\x_k = \x(t + k\Delta t)$, $\u_k = \u(t + k\Delta t) $, respectively.

\begin{algorithm}[t!]
\caption{HTMPC}\label{alg:HTMPC}
\begin{algorithmic}[1]
\Require \textit{Feasible sets} ($\X, \U$) \textit{, Initial State} $\x_o$, \textit{, Initial Guess} $\txb, \tub$  \textit{, and tracking cost functions} $\bar{\mathcal{J}}$
\State $\xb^0, \ub^0 \gets \txb, \tub$
\For{$l= 1, \dots, L$}
\State $\textit{NLP} \gets \textit{constructSTMPC}(\{\xb^\revision{i^*}, \ub^\revision{i^*}\}_{i=1}^{l-1},\x_o, \X, \U)$
\State  $\xb^\revision{l^*}, \ub^\revision{l^*} \gets \textit{SQPSolve}(\textit{NLP},\xb^{l-1}, \ub^{l-1}, \textit{MAX\_ITER})$
\EndFor
\State \Return $\xb^l, \ub^l$
\end{algorithmic}
\end{algorithm}

Similar to a \revision{typical} MPC controller, HTMPC has constraints for motion model \eqref{eq:HT-MPC-mdl}, collision avoidance \eqref{eq:HT-MPC-collisioncst}, state \eqref{eq:HT-MPC-xucst} and control input as well as state initialization \eqref{eq:HT-MPC-xocst}. As in \cite{gaertner_collision-free_2021}, we cover the robot with spheres and require the signed distance between theses spheres $\mathbf{sd}(\x_k)$ to be greater than a safe distance $\pmb{\delta}_\mathit{safe}$.  Instead of a scalar cost function, HTMPC has a vector of scalar cost functions \eqref{eq:HT-MPC-cost-vec}. Each scalar cost function represents the accumulated tracking error for a task in the sequence and is defined as
\begin{equation} \label{eq:HT-MPC-cost-scalar}
    \mathcal{J} = \frac{1}{2}\Big\{ \sum_{k=0}^{N-1} {\pnorm{\mathbf{e}_{k}}{\mathbf{Q}_k}}^2 + {\pnorm{\mathbf{e}_{N}}{\mathbf{P}}}^2\Big\},
\end{equation}
where matrices $\mathbf{Q}_k$ and $\mathbf{P}$ are diagonal positive semi-definite (PSD) weight matrices for stage and terminal cost with non-negative diagonal terms. 
Task index $l$ and dependency on $\x$ are dropped in \eqref{eq:HT-MPC-cost-scalar} for simplicity. \revision{HTMPC enforces the time-ordered sequence in the tasks by establishing a hierarchy in their cost functions. In particular,} the cost functions are arranged into a vector in the same order as in the task sequence\revision{; the cost function $\mathcal{J}_l$ corresponds to the $l^{th}$ task in the given list, $\task_{l_o-1+l}$, for which we will use $\task_\mathit{l}$ going forward.}
Lexicographic optimality requires that a task $\task_l$ is tackled only when optimal tracking has been established for preceding tasks; 
\revision{hence, the time-ordered sequence is maintained. }

\subsection{Single-Task MPC: A Building Block for Solving HTMPC}
We follow \autoref{alg:Lex-Opt} to solve the HTMPC problem \eqref{eq:HT-MPC}. In iteration $l$, the following scalar optimization problem is solved to optimize $\task_l$:
\begin{subequations}\label{eq:ST-MPC}
\begin{align}
      \min_{\bar{\x}, \bar{\u}} \; &\mathcal{J}_l + \mathcal{J}_\revision{\mathit{eff}} \\
 \text{s.t.} \, & \eqref{eq:HT-MPC-mdl}, \eqref{eq:HT-MPC-collisioncst}, \eqref{eq:HT-MPC-xucst}, \eqref{eq:HT-MPC-xocst}\\
 & {h}_i(\bar{\x}, \bar{\u};\bar{\x}^\revision{i^*}, \bar{\u}^\revision{i^*} ) \leqslant 0, \forall i=1, \cdots,l-1 \label{eq:ST-MPC-lexcst}\;.
\end{align}
\end{subequations} We denote the optimal state and control sequences to \eqref{eq:ST-MPC} as $\xb^\revision{l^*}, \ub^\revision{l^*}$. The parameters in \eqref{eq:ST-MPC-lexcst} $\xb^\revision{i^*}, \ub^\revision{i^*}$ are the optimal solution to $\task_i$ obtained in iteration $l=i$.

Compared to \eqref{eq:lex-opt-st}, we made two changes to the formulation. First, we added a control effort cost function to the objective:
\begin{equation}
    \mathcal{J}_\revision{\mathit{eff}} = \frac{1}{2} \Big \{\pnorm{\xb}{\bar {\mathbf{S}}}^2 + \pnorm{\ub}{\bar{\mathbf{R}}}^2  + \pnorm{\Delta\ub}{\bar{\mathbf{W}}}^2\Big\},
\end{equation}
where $\bar{\mathbf{S}}$, $\bar{\mathbf{R}}$ and $\bar{\mathbf{W}}$ are diagonal weight matrices for robot state $\xb$, control inputs $\ub$, and time difference of control inputs $\Delta\ub$ with non-negative terms of appropriate dimensions. Similar to other MPC methods, $\mathcal{J}_\revision{\mathit{eff}}$ encourages smooth motion and improves numerical stability. Similar to the HTIDKC method adding regularization terms, this could also improve numerical stability in case of singularity \cite{escande_hierarchical_2014}.

\revision{Second, we reformulated the lexicographic optimality constraint as \eqref{eq:ST-MPC-lexcst}, for which we consider two formulations}. \revision{The first} formulation closely resembles \eqref{eq:lex-opt-st-lexopt-constraint}:
\begin{equation} \label{eq:ST-MPC-lexcst-baseline}
    \mathcal{J}_i(\xb, \ub) \leqslant \mathcal{J}_i(\xb^\revision{i^*}, \ub^\revision{i^*}).
\end{equation}
However, the equality has been replaced with an inequality to reflect the fact that $\xb^\revision{i^*}, \ub^\revision{i^*}$ no longer minimizes the accumulated tracking error but the sum of both tracking error and control effort. 
\revision{Similar to \cite{romano_prioritized_2015}, we also proposed to approximate \eqref{eq:ST-MPC-lexcst-baseline} with decoupled constraints}
\begin{equation}\label{eq:ST-MPC-lexcst-box}
\vert {e_k}_p \vert \leqslant \vert{e_k}_p^\revision{i^*}\vert, \, \text{for}\, k=0, \dots, N, \,  p = 1, \dots, s ,
\end{equation}
directly limiting the tracking error $\mathbf{e}$ at prediction time step~$k$ along task space dimension $p$ with its optimal value $\mathbf{e}_k^\revision{i^*} = \mathbf{f}(\q_k^\revision{i^*}) - {\mathbf{r}}_k$. Note that the subscript $i$ for $\task_i$ is omitted in \eqref{eq:ST-MPC-lexcst-box} for simplicity.
The \revision{decoupled} constraints \eqref{eq:ST-MPC-lexcst-box} is an inner approximation of \eqref{eq:ST-MPC-lexcst-baseline}, which, we found, offers a better convergence rate when solving \eqref{eq:ST-MPC}.

\begin{figure}[t!]
      \centering
      \includegraphics[scale=1.0,trim={0 2.4cm 0 0}, clip]{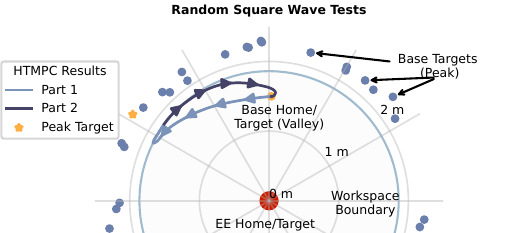}
      \caption{Illustration of 25 random square wave tests plotted in end effector frame using polar coordinates. The hierarchical tasks are, $\task_0$ minimizing constraints violation, $\task_1$ holding its end effector in home position, and $\task_2$ base tracking a square wave trajectory. The square wave trajectories have a duration of $16s$ that peaks during $t=0\sim8s$ (Part 1) and bottoms during $t=8\sim16s$ (Part 2). Valley target is the same for all cases which is the robot's home position. Peak targets are randomly selected outside the robot's workspace. An example base trajectory executed by HTMPC is provided with arrows indicating the direction of motion. Its corresponding peak target is marked as a yellow star. \revision{Note that \autoref{fig:radial_heatmap} is plotted in the same coordinate to show the correlation between the base tracking performance and the base targets' position.} }
      \label{fig:squarewave}
\end{figure}

\subsection{Solving HTMPC}
We present an algorithm for solving HTMPC in \autoref{alg:HTMPC}. Overall, we follow the sequential approach outlined in \autoref{alg:Lex-Opt} but without the uniqueness test. In each iteration, an \revision{Single-Task MPC (STMPC)} problem~\eqref{eq:ST-MPC} is constructed using the initial state, feasible sets for robot constraints, and solutions from previous iterations for lexicographic constraints. We choose to solve STMPC, a nonlinear program (NLP), using the Sequential Quadratic Programming (SQP) approach. SQP solves NLPs with a sequence of QPs via incremental quadratization with a maximum iteration $\textrm{MAX\_ITER}$. Each QP gives a local update direction along which the best step size is determined based on convergence and feasibility conditions. The initial guess for the SQP, $\xb^{l-1}$ and $\ub^{l-1}$, is either provided externally for the first iteration or solutions from the previous iteration.

Optimality of \autoref{alg:HTMPC} depends on the STMPC solutions. SQP approach finds a locally optimal solution to STMPC if converged \cite{boyd_convex_2004}. Therefore, the lexicographic optimality constraints \eqref{eq:ST-MPC-lexcst} only enforce lexicographic order in a local region. As a result, solutions by the proposed \autoref{alg:HTMPC} are locally optimal. If STMPC reaches $\text{MAX\_ITER}$ before converges, the incremental update is still an admissible step towards a local minima of \eqref{eq:HT-MPC} \cite{tazaki_constraint-based_2014}.

We share some notes on implementing HTMPC on a real system. The inequality constraints in STMPC are formulated as hard constraints. However, when running on a real system, disturbances, sensor noises, and state estimation errors might render the MPC problem infeasible, leading to a complete failure if no remedies are taken. In recent years, it has become more common to soften these constraints in implementation. One method is to incorporate them into cost functions via barrier functions. It is also possible to relax the inequality constraints with slack variables whose norm is minimized in the objective. The trade-off between minimizing the original cost function and constraints violation is controlled indirectly by tuning weight parameters. We used the relaxed log barrier function method for state and control constraints \eqref{eq:HT-MPC-xucst} as well as collision constraints \eqref{eq:HT-MPC-collisioncst}. 

For the lexicographic optimality constraints \eqref{eq:ST-MPC-lexcst}, we took an alternative approach to have quantitative control over their relaxation. More specifically, in the line search step of SQP, instead of finding a step size that strictly satisfies the lexicographic constraints, we allow violations within a predefined range. That is, a step size is feasible if it satisfies
\begin{equation}\label{eq:ST-MPC-lexcst-relaxation}
    {h}_i \leqslant \delta_i, \; \forall i=1, \cdots,l-1 ,
\end{equation}
where $\delta_i \in \Real >0 $ is the tolerance for $\task_i$. Equation~\eqref{eq:ST-MPC-lexcst-relaxation} allows us to quantify the maximal tracking error that preceding tasks can compromise to improve subsequent tasks at each control time step. 
This becomes useful in practice where different tasks and applications require different levels of control accuracy. In the next section, we will show that this can be leveraged to improve hierarchical task performance.

\section{Experiments}

 \begin{figure}[t!]
      \centering
      \includegraphics[trim={0.1cm 0.1cm 0 0.cm}, clip]{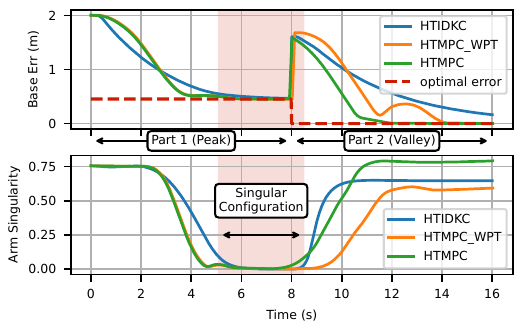}

      \caption{Experimental results of the three competing methods for the example shown in \autoref{fig:squarewave}. Since the peak target is not reachable, all three methods converge to a non-zero base error in Part 1 when the robot arm becomes fully extended and singular. For Part 2, only the proposed methods HTMPC and HTMPC\_WPT are converged to its valley target. HTIDKC did not converge because the parameters were optimized for Part 1.}
      \label{fig:htmpc_vs_htidkc_error_trajectory}
\end{figure}

Our experimental platform is a 9 DoF mobile manipulator platform consisting of a holonomic Ridgeback mobile base and a 6 DoF UR10 arm (\autoref{fig:main}). 
\revision{The pose of the mobile base is measured by an external motion tracking system, whereas the EE pose is computed from the arm’s joint positions measured by its internal joint encoders.}
The proposed HTMPC is implemented using the \textit{CasAdi} framework in Python \cite{andersson_casadi_2019}. A standard SQP procedure in \cite{boyd_convex_2004} with the proposed back-tracking line search was implemented for STMPC with QP solver from \textit{Gurobi} \cite{gurobi_optimization_llc_gurobi_2023}. For all results in this section, we used $\Delta t = 0.1s$, $N = 10$, $\textrm{MAX\_ITER} = 1$. The average compute time of HTMPC is 63ms, so its control frequency was set at 10Hz.  
All results are experimental data except for those in \autoref{sec:results-lex}. Videos of all experimental results can be found at \href{http://tiny.cc/htmpc}{http://tiny.cc/htmpc}.


\subsection{Hierarchical Random Square Wave Tests} \label{sec:results-test}
\revision{To see how well HTMPC enforces the time-ordered sequence with hierarchy, we evaluated it with 25 randomly generated tests with three-level hierarchical tasks illustrated in \autoref{fig:squarewave}}. \revision{The two tracking tasks represent the sub-sequence given to the controller as part of the process that fulfills the high-level user request.}
The test is designed to see how controllers can handle task changes, singular configurations, and variations in reference trajectories. 
We chose square wave trajectories for the base \revision{to test the controller's capability in reacting to changes in the subsequent task initiated by the upper-level planning module.  } 

An example base trajectory executed by HTMPC is shown in \autoref{fig:squarewave} as two parts. The base moves from its home position to reach the peak target (yellow star) (Part 1) before moving back to its home position (Part 2). The base only moves as far as its workspace boundary to avoid compromising the EE task. The corresponding tracking error is shown in \autoref{fig:htmpc_vs_htidkc_error_trajectory}. In Part 1, the base tracking error converges to its optimal steady-state error as it comes to its workspace boundary with the arm fully extended. In Part 2, the base tracking error should converge to zero since the valley target is reachable.

\subsection{Lexicographic Optimality Constraints}\label{sec:results-lex}

\begin{table}[t]
\caption{HTMPC EE Tracking Error with Different Lexicographic Optimality Constraints (mm)}\label{tbl:lex} 
\begin{center}
\begin{tabular}{|c||c|c|c|c|}
\hline
\multirow{2}{*}{Eq} &\multicolumn{2}{|c|}{Low $\delta$}  &\multicolumn{2}{|c|}{High $\delta$}\\\cline{2-5}
&avg &std &avg &std\\
\hline
\revision{Baseline \eqref{eq:ST-MPC-lexcst-baseline} } &3.24 &0.56 &6.25 &2.73 \\\cline{1-5}
  \revision{Proposed \eqref{eq:ST-MPC-lexcst-box}} &\textbf{3.07} &\textbf{0.73}&\textbf{3.42 }&\textbf{1.55}\\
 \hline
\end{tabular}
\end{center}
\end{table}

\begin{figure}[t!]
	\centering
	\includegraphics[trim={0.cm 0.1cm 0 0.cm}, clip]{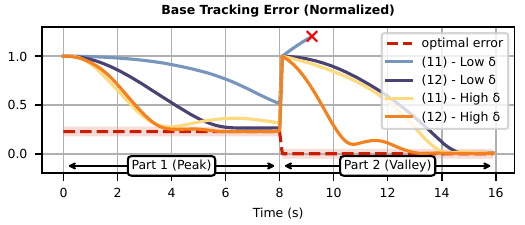}
	\caption{\revision{HTMPC normalized base tracking error with different lexicographic constraints. The red cross indicates that the base tracking has diverged and is removed from the plot. A desired steady state error bound of 5\% is shaded in red. Formulation \eqref{eq:ST-MPC-lexcst-box} outperforms \eqref{eq:ST-MPC-lexcst-baseline} with shorter convergence time for both tolerance levels. Increasing the constraint tolerance improves base tracking performance at the cost of the EE task (\autoref{tbl:lex}). }}
	\label{fig:base-err-lex}
\end{figure}

We tested the two lexicographic constraint formulations on the example square wave test shown in \autoref{fig:squarewave} \revision{and presented the results in \autoref{tbl:lex} and \autoref{fig:base-err-lex}}. The proposed relaxation method \eqref{eq:ST-MPC-lexcst-relaxation} was also implemented, and two levels of tolerance were tested. The tolerance value $\revision{\delta}$ was set at a comparable level for both formulations ($1cm$ EE tracking error for High $\revision{\delta}$, and $1mm$ for Low $\revision{\delta}$)
\footnote{Note that both formulations with low tolerance failed to maintain $e_{EE}$ within the desired error $1mm$ error bound. The main reason is that lexicographic constraints are imposed on the open-loop predicted trajectory. Therefore, closed-loop tracking error is not guaranteed.}. 
Our proposed base tracking error is reported separately for Part 1 and Part 2 \revision{in \autoref{fig:base-err-lex}}. Both are normalized by the initial error of each part. \revision{Our proposed formulation \eqref{eq:ST-MPC-lexcst-box}} leads to improved tracking performance for both tasks mainly because 
\revision{it better approximates the nonlinear feasible set after linearization, }permitting larger step sizes during line search. For both formulations, increasing tolerance can improve the secondary task but at the cost of the primary task. However, compromise is transient, so we only see a non-negligible increase to $e_\mathit{EE}$ standard deviation. 

\subsection{HTMPC vs HTIDKC}\label{sec:results-idkc}
We compared the tracking performance of HTMPC with HTIDKC using the 25 randomly sampled tests. 
HTIDKC is implemented following formulations in \cite{escande_hierarchical_2014}. \revision{The three-level tasks were precisely the same for both approaches with} the only task difference \revision{being} that an additional jerk bound was implemented for HTMPC to encourage smooth motions when running on a real robot, which was not necessary for HTIDKC. Regularization was also implemented in cost function for HTIDKC to deal with singularity. We used the cascaded QP approach with regularization terms included for singularity \cite{escande_hierarchical_2014} to solve HTIDKC and implemented it in \textit{CasAdi} with \textit{Gurobi} QP solver. The compute time for HTIDKC is 17ms on average, so its control frequency was set at 50Hz. To simplify lexicographic comparisons, the two controllers were tuned to have similar performance for the top two tasks and Part 1 for the base tracking task; therefore, we only need to compare Part 2 to determine their lexicographic optimality. 

In addition to HTIDKC, we also compared with a second baseline, HTMPC\_WPT. This method is HTMPC receiving the square wave trajectory as two separate waypoints instead of one trajectory. As a result, similar to HTIDKC, HTMPC\_WPT does not foresee the task change at t=8s.

\begin{figure}[t!]
	\centering
	\includegraphics[trim={0.2cm 0.15cm 0 0.cm}, clip]{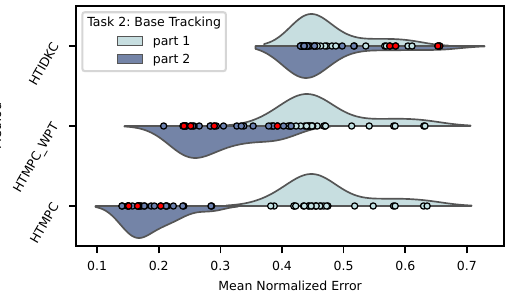}
	\caption{Violin plot showing the hierarchical task results of the three competing methods for the 20 random square wave tests. Both distributions (shaded area) and individual data points (dots) are shown. Results are presented separately for Part 1 (light blue) and Part 2 (dark blue). \revision{Part 2 results for the singularity impacted cases (\autoref{fig:radial_heatmap}) are highlighted (red)}. Parameters are designed so that all methods have similar performance for $\task_0$, $\task_1$ and Part 1 of $\task_2$. HTMPC achieves the best control performance in Part 2 of $\task_2$ with a $42\%$ improvement over the HTIDKC approach. 
  Therefore, HTMPC 
  \revision{achieves a lower cost of the hierarchical tracking problem} in the sense of lexicographic order as per \autoref{def:lex-opt-opt-def}.}
	\label{fig:htmpc_vs_htidkc_error_violin}
\end{figure}

Tracking error is plotted over control time in \autoref{fig:htmpc_vs_htidkc_error_trajectory} for the example in \autoref{fig:squarewave}. A violin plot showing a fitted distribution for the normalized base error is shown in \autoref{fig:htmpc_vs_htidkc_error_violin}. 
Part~1 is designed to have a similar performance. 
For Part~2, HTIDKC has the worst tracking performance of all three methods---it does not have any improvements over Part~1 despite the fact that the valley target is within reach and a lower mean error can be achieved. In contrast, both HTMPC methods have a lower tracking error for Part~2 mainly because the controller parameters were optimized for Part~1 and HTIDKC does not generalize to Part 2 as well as the optimization-based HTMPC controller. Parameter scheduling is required for HTIDKC to have comparable results to HTMPC. HTMPC also slightly outperforms HTMPC\_WPT, which can be attributed to the predictive nature of MPC. When receiving one complete trajectory instead of two separate waypoints, HTMPC foresees the coming step in the trajectory and reacts earlier to have an overall lower tracking error. As seen in \autoref{fig:htmpc_vs_htidkc_error_trajectory}, HTMPC converges faster than HTMPC\_WPT in Part~2 while incurring negligible error in Part~1.

\begin{figure}[t!]
      \centering
      \includegraphics[trim={0 1.4cm 0 0.8cm}, clip]{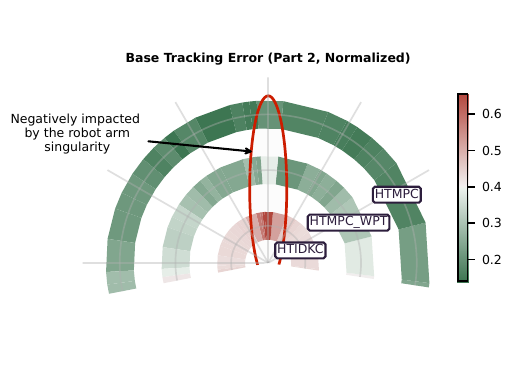}
      \caption{Radial heat map showing the spatial correlation between the normalized \revision{Part 2} base tracking error and peak targets. As in \autoref{fig:squarewave}, this plot is in polar coordinates where the angular component is the relative direction of base peak target to the end effector home position. Circled in red is the region where singularity plays a major role in tracking performance. HTMPC outperforms the other two methods in dealing with singularity.}
      \label{fig:radial_heatmap}
\end{figure}

Normalized base tracking error \revision{(Part 2)} is shown as a radial heat map in \autoref{fig:radial_heatmap} to illustrate its spatial correlation with peak targets. 
All three methods deteriorate as the base peak target, valley target and EE target become colinear \revision{in x-y plane (highlighted in red in both \autoref{fig:htmpc_vs_htidkc_error_violin} and \autoref{fig:radial_heatmap})}. 
\revision{In these cases, the optimal base motion demands aggressive robot arm maneuvers with high joint velocities to prioritize the EE task when the robot arm is near its singular configuration.} Of all three methods, HTMPC has the least decline in performance \revision{due to better regularization of both the joint velocity and acceleration and better constraint handling via prediction and line search.} 

\subsection{Hierarchical-Task vs Baseline Control Architectures}\label{sec:results-arch}
In this section, we compare the optimality of our proposed  \revision{HTMPC-based hierarchical-task control architecture (HT-Arch)} with a typical single-task architecture (\revision{ST-Arch}) as a baseline, \revision{the arm-base decoupled architecture similar to \cite{carius_deployment_2018}. The task sequence were given to the controller as described in \autoref{sec:method}, and we chose $L=1$ for ST-Arch and $L=2$ for HT-Arch. Although we only implemented the single-task method, the multi-task architecture \cite{burgess-limerick_architecture_2023} would have given similar results for the first task.}

\begin{figure}[t!]
      \centering
      \includegraphics[width=\linewidth, trim={0 0.05cm 0 0.cm}, clip]{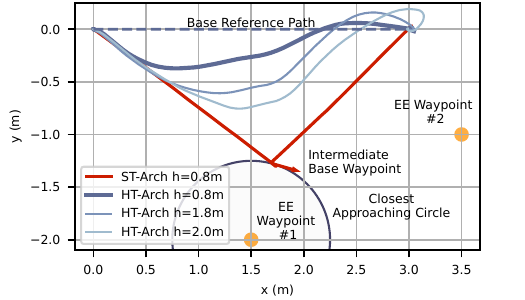}
      \caption{Mobile manipulator base path comparison of \revision{a typical single-task (ST-Arch) and our proposed hierarchical-task (HT-Arch)} architectures for a sequence of two EE waypoint tasks. The EE waypoint $\#1$ were set at different heights. The intermediate base waypoint for the \revision{ST-Arch} is determined following \cite{burgess-limerick_architecture_2023}. \revision{Compared to ST-Arch, the proposed HT-Arch approach} (\revision{HT-Arch} h=0.8m) optimally positions the robot base where the EE waypoint $\#1$ just becomes reachable.}
      \label{fig:arch_path_comparison_1}
\end{figure}

The first task is illustrated in \autoref{fig:arch_path_comparison_1}, where the robot needs to visit two EE waypoints in a sequence. Although EE waypoints are the same for both approaches, base reference paths \revision{were optimized differently for} the underlying controller. With a decoupled arm and base controller, base reference path for \revision{ST-Arch} needs to go through an intermediate waypoint at a predefined distance from the first EE target before reaching the second one. In contrast, since HTMPC can perform redundancy resolution using robot's kinematics, it is sufficient to have a path going straight towards the second EE target \revision{for HT-Arch}. Actual base paths executed by the controllers are also different. \revision{ST-Arch} 
closely follows the given reference to reach the EE target whereas \revision{HT-Arch} deviates to optimally position its base while holding its EE on target. Additionally, HTMPC \revision{in HT-Arch} can automatically adapt its base path to EE targets at different heights, whereas \revision{ST-Arch relies on the motion planner to adjust the intermediate base waypoint for its controller based on heuristics \cite{burgess-limerick_enabling_2023}}. 


\begin{figure*}
        \centering
    \includegraphics[width=\textwidth, trim={0 0.75cm 0.05cm 0.2cm}, clip]{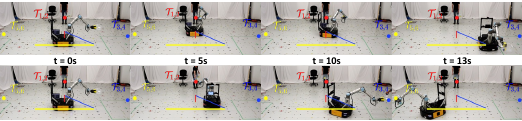}
\caption{Robot performing a sequence of delivery tasks to three people. \textit{Top figure} shows our robot running an arm-base decoupled single-task control architecture. \textit{Bottom figure} shows robot running a our proposed HTMPC control architecture. Our proposed architecture is more efficient in executing sequential mobile manipulation tasks by allowing the base to move its next delivery target while holding EE in place.}
\label{fig:results-demo-htmpc}
\end{figure*}

We also compared the two architectures on time efficiency with a long horizon task where the robot needs to perform a sequence of delivery tasks to three people for four rounds. In this test, both the \revision{decoupled arm and base} controller and HTMPC were given the same base and EE plans. Leveraging HTMPC's multi-tasking nature for sequential tasks, \revision{our proposed HT-Arch approach} executes the tasks 2.3 times faster than the \revision{ST-Arch} approach (\autoref{fig:results-demo-htmpc}).

We also introduced changes to the robot delivery task to demonstrate the task-space reactivity of our proposed architecture. In the first scenario, the robot is forced to hold its EE in place until the person picked out the ball whose timing is unknown to the controller. HTMPC allowed the base to continue moving toward its next target as usual but stopped the base immediately before compromising the EE task. Moreover, HTMPC automatically updated the base position as the next delivery target changes location. We also showed that our proposed architecture can quickly react to an unexpected shift in the current task via fast replanning and efficient hierarchical control. Both tests can be found in the accompanied video and also at \href{http://tiny.cc/htmpc}{http://tiny.cc/htmpc}.

\section{Conclusions}
We presented the HTMPC framework for solving sequential mobile manipulation tasks. \revision{The framework centers around the HTMPC controller that enforces the time-ordered sequence of tasks with a hierarchy in their trajectory tracking cost functions. We demonstrated that} HTMPC outperforms the state-of-the-art HTIDKC
approach in hierarchical trajectory tracking tasks. \revision{We also presented a hierarchial-task control architecture optimized for HTMPC to leverage the robot's redundancy for sequential tasks.} \revision{We demonstrated that our proposed architecture has improved efficiency and reactivity in executing sequential tasks compared to the existing architectures}. \revision{In future works, we wish to investigate alternative regularization and lexicographic constraint formulations to improve our proposed HTMPC algorithm.}

\bibliographystyle{IEEEtran}
		\bibliography{sections/references}

\addtolength{\textheight}{-12cm}   

\end{document}